\documentclass[runningheads,a4paper]{llncs}

\usepackage{amssymb}
\usepackage{graphicx}
\usepackage{color}
\usepackage{multirow}
\usepackage{pgffor}
\usepackage{subfig}

\usepackage{url}

\begin{document}

\mainmatter  

\title{Generative Adversarial Networks 
for\\ MR-CT Deformable Image Registration}

\titlerunning{GANs for MR-CT Deformable Image Registration}

%
%
\author{Christine Tanner \and  Firat Ozdemir 
\and Romy Profanter  \and Valeriy Vishnevsky \and \\ Ender Konukoglu \and Orcun Goksel}
\authorrunning{Tanner et al.}

\institute{Computer Vision Lab, ETH Zurich,
8092 Zurich, Switzerland\\
}
%
%

\maketitle

\begin{abstract}
Deformable Image Registration (DIR) of MR and CT images is one of the most challenging registration task, due to the inherent structural differences of the modalities and the missing dense ground truth. Recently cycle Generative Adversarial Networks (cycle-GANs) have been used to learn the intensity relationship between these 2 modalities for unpaired brain data. Yet its usefulness for DIR was not assessed.

In this study we evaluate the DIR performance for thoracic and abdominal organs after synthesis by cycle-GAN. We show that geometric changes, which differentiate the two populations (e.g. inhale vs. exhale), are readily synthesized as well. This causes substantial problems for any application which relies on spatial correspondences being preserved between the real and the synthesized image (e.g. plan, segmentation, landmark propagation). To alleviate this problem, we investigated reducing the spatial information provided to the discriminator by decreasing the size of its receptive fields.

Image synthesis was learned from 17 unpaired subjects per modality.
Registration performance was evaluated with respect to manual  segmentations of 11 structures for 3 subjects from the VISERAL challenge. State-of-the-art DIR methods based on Normalized Mutual Information (NMI), Modality Independent Neighborhood Descriptor (MIND) and their novel combination achieved a mean segmentation overlap ratio of 76.7,  67.7, 76.9\%, respectively. This dropped to 69.1\% or less when registering images synthesized by cycle-GAN based on local correlation, due to the poor performance on the thoracic region, where large lung volume changes were synthesized.
Performance for the abdominal region was similar to that of CT-MRI NMI registration (77.4 vs. 78.8\%) when using 3D synthesizing MRIs (12 slices) and medium sized receptive fields for the discriminator. 
\end{abstract}

\section{Introduction}

Deformable Image Registration (DIR) is a challenging task and active field of research in medical image analysis~\cite{Sotiras2013}. 
Its main application is fusion of the information acquired by the different modalities to facilitate diagnosis and treatment planning~\cite{Sotiras2013}. 
For example, in radiotherapy treatment planning Magnet Resonance (MR) images are used to segment the tumor and organs at risk, while the tissue density information provided by the corresponding Computer Tomography (CT) image is used for dose planning~\cite{wolterink2017deep}. 
CT and MR images are acquired using separate devices and often on different days.
Therefore the patient will not be in exactly the same posture and the position of inner organs might change, due to respiration, peristalsis, bladder filling, gravity, etc. Thus, DIR is needed.
The main difficulty of MR-CT DIR is the definition of an image similarity measure, which reliably quantifies the local image alignment for optimizing the many free parameters of the spatial transformation.  
This is an inherent problem as multi-modal images are acquired because they provide complementary information.

\noindent
{\bf Multi-modal similarity measures. }
The main voxel-wise multi-modal image (dis)similarity measures 
are (i) statistical measures that use intensity information directly and try to maximize (non-linear) statistical dependencies between the intensities of the images (e.g. Normalized Mutual Information (NMI)~\cite{Studholme1999}, MI~\cite{Pluim2000}), and (ii) structural measures based on structural representations that try to be invariant to different modalities (e.g. normalized gradient fields~\cite{Haber2006}, entropy images~\cite{Wachinger2012}, Modality Independent Neighborhood Descriptor (MIND)~\cite{Heinrich2012}).

\noindent
{\bf Intensity remapping. }
The drawback of structural representations is that all unstructured (e.g. homogenous) regions are mapped to the same representation regardless of their original intensity. To avoid this information reduction, methods to directly re-map intensities have been proposed~\cite{Andronache2008,Wachinger2012,Knops2004}. 
The joint histogram of the coarsely registered images was employed to remap the intensities of both images to a common modality based on the least conditional variance to remove structures not visible in both images~\cite{Andronache2008}.
Assuming that the global self-similarity of the images (i.e. the similarities between all image patches) is preserved across modalities, intensities were mapped into a 1D Laplacien Eigenmap based on patch intensity similarity~\cite{Wachinger2012}.
A k-means clustering based binning scheme, to remap spatially unconnected components with similar intensities to distinct intensities, was proposed in~\cite{Knops2004} for retina images. 
While these intensity-remappings provide some improvements, they are simplifications to the underlying complex relationship between the intensity of the two modalities.  

\noindent
{\bf Learning from paired data. }
Given aligned multi-modal training data, attempts have been made to learn this complex relationship.
The last layer of a deep neural network (DNN) classifier, which discriminates between matching and not matching patches, was used to directly learn the similarity measure~\cite{Cheng2016}. The DNN was initialized by a stacked denoised autoencoder, where the lower layers were separately trained per modality to get modality-dependent features.
It was observed that 
the learned CT filters look mostly like edge-detectors, while the MR filters detect more complex texture features.
In~\cite{So2017} the expected joint intensity distribution was learned from co-registered images. The dissimilarity measure was then based on the Bhattacharyya distance between the expected and observed distribution.
Machine learning has been used to learn how to map one modality to the other. \cite{Roy2014} synthesized CT from MR brain images
by matching MR patches to an atlas (created from co-registered MR and CT images) and augmented these by considering all convex patch combinations.
\cite{Cao2017} proposed a bi-directional image synthesis approach for non-rigid registration of the pelvic area, where random forests are trained on Haar-like features extracted from pairs of pre-aligned CT and MR patches. An auto-context model was used to incorporate neighboring prediction results. 
All these learning-based approaches depend on co-registered multi-modal images for training. This is very difficult for deforming structures as dense (voxel-wise) spatial correspondences are required and 
CT and MR images cannot be acquired simultaneously yet~\cite{Liu2017}.

\noindent
{\bf Learning without paired data. }
A cross-modality synthesis method which does not require paired data was proposed in~\cite{vemulapalli2015unsupervised}.
It is based on generating  multiple target modality candidate values for each source voxel independently using cross-modal nearest neighbor search. A global solution is then found by simultaneously maximizing global MI and local spatial consistency. Finally, a coupled sparse representation was used to further refine the synthesized images. 
When applied to T1/T2 brain MRIs, T1 images were better synthesized than T2 images (0.93 vs. 0.85 correlation to ground truth). Extending the method to a supervised setting outperformed state-of-the-art supervised methods slightly.

Recently cyclic-consistent Generative Adversarial Networks (cycle-GANs) were proposed for learning an image-to-image mapping between two domains ($\mathcal{A}$\&$\mathcal{B}$) from unpaired datasets~\cite{zhu2017unpaired}. The method is based on two generator networks ($G_{\mathrm{B}}$ to synthesize image $\hat{\mathbf{I}}_\mathrm{B}$ from $\mathbf{I}_\mathrm{A}$, $G_\mathrm{A}$) and two discriminator networks ($D_\mathrm{A}$, $D_\mathrm{B}$). Besides the usual discriminator loss to differentiate synthesized and real images (e.g. $\hat{\mathbf{I}}_\mathrm{A}, \mathbf{I}_\mathrm{A}$), a cycle loss was introduced which measures the difference between the real image and its twice synthesized image, e.g. $|\mathbf{I}_\mathrm{A} - G_\mathrm{A}(G_\mathrm{B}(\mathbf{I}_\mathrm{A}))|_1$. Good performances were shown for various domain translation tasks like labels to photos and arial photos to maps. Very recently, this promising approach was employed for slice-wise synthesizing CT from MR head images from unpaired data~\cite{wolterink2017deep}. It achieved lower mean squared errors (74 vs. 89 HU) than when training the same generator network on rigidly aligned MR and CT data~\cite{nie2017medical}. It was reasoned that this could be due to misalignments, as the images contained also deforming structures (e.g. neck, mouth).
Cycle-GANs were used for synthesis of MR from unpaired CT images for enriching a cardiac dataset for training  thereafter a segmentation network~\cite{chartsias2017adversarial}. A view alignment step using the segmentations was incorporated to make the layout (e.g. position, size of anatomy) of the CT and MR images similar, such that the discriminator cannot use the layout to differentiate between them. Furthermore the myocardium mask for both modalities was provided during training, as the cycle-GAN not only changed the intensities but also anatomical locations such that the mask was no longer in correspondence with the image. Hence this is  not a completely unsupervised approach.
Similarly, a shape-consistency loss from segmentations was incorporated in~\cite{zhang2018translating} to avoid geometric changes between the real and synthesized images. It was argued that "from the discriminator perspective, geometric transformations do not change the realness of synthesized images since the shape of training data is arbitrary". However this does not hold if there is a geometric bias between the two datasets.

Synthesized MR PD/T1 brain images via 
patch matching were shown to be useful for segmentation and inter-modality cross-subject registration~\cite{iglesias2013synthesizing}. If this also holds for MR-CT synthesis via cycle-GANs for thoracic and abdominal regions has not yet been studied.
Our contributions include 
(i) combining two state-of-the-art multi-modal DIR similarity measures (NMI, MIND), (ii) studying the effect of the image region size on the consistency of the synthesized 3D images, and (iii) evaluating the usefulness of synthesized images for deformable registration of CT and MR images from the thorax and abdomen against a strong baseline.

\section{Materials}

We used 17 unpaired and 3 paired 3D MR-CT images from the VISCERAL Anatomy3 benchmark training set (unenhanced, whole body, MR-T1) and their gold standard segmentations for evaluation~\cite{Jimenez-del-Toro2016}. 
The 3 subjects with paired data had IDs 10000021, 10000067 and 10000080. 
All MRIs were bias field corrected 
using the N4ITK method~\cite{Tustison2010}. 
All images were resampled to an isotropic resolution of 1.25~mm.
This was motivated by the image resolution of the original MRIs being 1.25$\times$6$\times$1.25~mm$^3$ in left-right, posterior-anterior and superior-inferior direction. The CT images had a resolution between 0.8$\times$0.8$\times$1.5~mm$^3$ and 1.0$\times$1.0$\times$1.5~mm$^3$. 

To reduce memory requirements, we automatically extracted from each image two regions such that the included gold standard segmentations were at least 5~mm away from the inferior and superior region boundary. The thorax region covered the segmentations of the liver, spleen, gallbladder, and right and left lung. The abdominal region contained the bladder, lumbar vertebra 1, right and left kidney, and right and left psoas major muscle, see Figs.~\ref{fig:fakeCT},~\ref{fig:fakeMR}, left column.

Closer investigation of poor performing registration results showed that for case 10000067 the segmentation labels of the right and left kidney were swapped in the MRI. Additionally, for 10000080 the segmentation of the lumbar vertebra 1 in the MRI seems to be that of lumbar vertebra 2. 
We corrected these kidney annotations and excluded this lumbar vertebra 1 segmentations from the results.

\section{Method}

\subsection{Image Synthesis}

{\bf Cycle-GAN. } For image synthesis, 
we followed the cycle-GAN network architecture as described in~\cite{zhu2017unpaired,wolterink2017deep}, starting from an existing implementation\footnote{https://github.com/xhujoy/CycleGAN-tensorflow}. 
In short, the two generators ($G_\mathrm{CT}$, $G_\mathrm{MR}$) are 2D fully convolutional networks with 9 residual blocks and two fractionally strided convolution layers (res-net).
The discriminators ($D_\mathrm{CT}$, $D_\mathrm{MR}$) are fully convolutional architectures to classify overlapping $P$$\times$$P$ image patches as real or fake (PatchGAN)~\cite{isola2016image}\footnote{The discriminators consist of 5 convolutions layers (I256-C128-C64-C32-C32, stride length 2-2-2-1-1, 4$\times$4 kernels) for $P$$=$$70$ and 4 layers (I256-C128-C64-C64, stride length 2-2-1-1) for $P$$=$$34$. Leaky ReLU activation functions (factor 0.2) were used. Data was normalized by instance normalization.}.

The networks take input images of size 256$\times$256 pixels and $C$ channels.  
Larger-sized test images were synthesized from the average result of 256$\times$256$\times$$C$ regions extracted with a stride length of $S$$\times$$S$$\times$$S_C$. 
The cycle-GAN was optimized to reduce the overall loss $L$,
which is a weighted sum of the discriminator losses $L_{\mathrm{CT}}$, $L_\mathrm{MR}$ and the generator cyclic loss $L_{\mathrm{cyc}}$:
\begin{eqnarray}
L&=&L_{\mathrm{CT}}+L_{MR}+\lambda_{\mathrm{cyc}} L_{\mathrm{cyc}}\\
L_{\mathrm{CT}}&=&(1 - D_{\mathrm{CT}}(I_{\mathrm{CT}}))^2 + \label{eq:disCT}
D_{\mathrm{CT}}(G_{\mathrm{CT}}(I_\mathrm{MR} ))^2\\
\label{eq:disMR}
L_\mathrm{MR}&=&(1 - D_\mathrm{MR}(I_{MR}))^2 + D_\mathrm{MR}(G_\mathrm{MR}(I_{\mathrm{CT}} ))^2\\
L_{\mathrm{cyc}}&=&||G_{\mathrm{CT}}(G_\mathrm{MR}(I_{\mathrm{CT}}))-I_{\mathrm{CT}}||_{1} + 
||G_\mathrm{MR}(G_{\mathrm{CT}}(I_\mathrm{MR}))-I_\mathrm{MR}||_{1}
\end{eqnarray}

\subsection{Image Registration}

{\bf Rigid Registration. } The CT and MR images were first rigidly registered using the function \emph{imregister} from the MATLAB Image Processing Toolbox~\cite{DocumentationMatlabImregister}, set to multi-modal configuration (Mattes Mutual Information, one-plus-one evolutionary optimization). These rigid registration results were then used as starting points for all subsequent deformable image registrations.

\noindent
{\bf Deformable Registration - MIND. }
The so-called modality independent neighborhood descriptor (MIND) was proposed as dissimilarity measure for multi-modal DIR~\cite{Heinrich2012}.  MIND is based on a multi-dimensional descriptor $\mathbf{s}_\mathrm{MIND}$ per voxel $\mathbf{x}$, which captures the self-similarity of the image patch around $\mathbf{x}$ (denoted as $\mathbf{P}(\mathbf{x})$) with the patches $\mathbf{P}(\mathbf{x}+\mathbf{r})$ 
in a local neighborhood  $\mathcal{N}$ of $\mathbf{x}$. 
The single entries $\mathbf{s}_\mathrm{MIND}(\mathbf{I},\mathbf{x},\mathbf{r}_i)$ are calculated by a Gaussian function
\begin{equation}
\mathbf{s}_\mathrm{MIND}(\mathbf{I},\mathbf{x},\mathbf{r}_i) = \frac{1}{n} \, \exp \left(-\frac{d_\mathrm{p}(\mathbf{I},\mathbf{x},\mathbf{r}_i)}{v(\mathbf{I},\mathbf{x})} \right)
\end{equation}
where $n$ is a normalization constant such that the maximum value in $\mathbf{s}_\mathrm{MIND}$ is 1,  $d_\mathrm{p}$ defines the patch dissimilarity $d_\mathrm{p}(\mathbf{I},\mathbf{x},\mathbf{r}) $$=$$ \sum_{\mathbf{x}_j \in \mathbf{P}(\mathbf{x})} \mathbf{G}_{\sigma}(\mathbf{x}_j)  (\mathbf{P}(\mathbf{x}_j)-\mathbf{P}(\mathbf{x}_j+\mathbf{r}))^{2}$ with Gaussian kernel $\mathbf{G}_{\sigma}$ of the same size as patch $\mathbf{P}(\mathbf{x})$ and the half-size of the patch being equal to $\lceil{1.5\sigma}\rceil$. 
$v$ is the variance of a six-neighborhood search region.
$\mathbf{s}_\mathrm{MIND}$ is calculated in a dense fashion for each image independently. 
The dissimilarity $E_\mathrm{MIND}(\mathbf{A},\mathbf{B})$ of images $\mathbf{A}$ and $\mathbf{B}$ is finally defined by
$E_\mathrm{MIND}(\mathbf{A},\mathbf{B})$$ = $$\sum_{\mathbf{x} \in \Omega} E_\mathrm{MIND}(\mathbf{A},\mathbf{B},\mathbf{x})^{2}$ with
\begin{eqnarray}
E_\mathrm{MIND}(\mathbf{A},\mathbf{B},\mathbf{x}) & = & \frac{1}{|\mathcal{N}|} \sum_{\mathbf{r}_i \in \mathcal{N}} | \mathbf{s}_\mathrm{MIND}(\mathbf{A}, \mathbf{x}, \mathbf{r}_i) - \mathbf{s}_\mathrm{MIND}(\mathbf{B},\mathbf{x},\mathbf{r}_i)|.
\end{eqnarray}
In the MIND registration framework, the images are downsampled via Gaussian pyramids and the deformation field is regularized via the squared L2-norm.   
Additionally after each Gauss-Newton update step during optimization,
each deformation field is replaced by combining half of its own transformation with half of the inverse transformation of the other deformation field (see~\cite{Heinrich2012,Avants2008}) to obtain diffeomorphic transformations.
We used the provided code~\cite{HeinrichCode} and compared results after integrating the MIND measure in our DIR method~\cite{vishnevskiy2017isotropic}.

\noindent
{\bf Deformable Registration - ourDIR. }
We extended our DIR method, based a linearly interpolated grid of control points and various displacement regularization measures,
to incorporate the multi-modal (dis)similarity measures normalized mutual information (NMI) and MIND, and their combination NMI+MIND. 

Given fixed image $\mathbf{I}_\mathrm{f}$, moving image $\mathbf{I}_\mathrm{m}$, displacements $\mathbf{k}$ at the control points, and interpolation function $d$ to get dense displacements,
the NMI dissimilarity 
$E_\mathrm{NMI}(\mathbf{I}_\mathrm{f},\mathbf{I}_\mathrm{m}(d(\mathbf{k})))$ is defined by
$E_\mathrm{NMI}(\mathbf{A,B})$$ =$$- (H_\mathbf{A}+H_\mathbf{B})/H_\mathbf{A,B}$, with marginal entropies $H_\mathbf{A}$, $H_\mathbf{B}$ and joint entropy $H_\mathbf{A,B}$ computed from intensity histograms with 100 equally-spaced bins between the 0.5 and 99.5 percentiles of the image intensity.
The gradients of $E_\mathrm{NMI}(\mathbf{I}_\mathrm{f},\mathbf{I}_\mathrm{m}(d(\mathbf{k})))$ with respect to $d(\mathbf{k})^{(i)}[x]$ are calculated as described in~\cite{Crum2003}.
To avoid infinity gradients, we replace zero probabilities with $1/(2N_\mathrm{V})$, where $N_\mathrm{V}$ is the number of image voxels.
We combined the dissimilarities NMI and MIND by
\begin{equation}
E_\mathrm{N+M}(\mathbf{I}_\mathrm{f},\mathbf{I}_\mathrm{m}(d(\mathbf{k}))) 
=
\beta E_\mathrm{NMI}(\mathbf{I}_\mathrm{f},\mathbf{I}_\mathrm{m}(d(\mathbf{k}))) 
+
(1-\beta) s E_\mathrm{MIND}(\mathbf{I}_\mathrm{f},\mathbf{I}_\mathrm{m}(d(\mathbf{k}))) 
\end{equation}
where $s$ is a scaling parameter to get $E_\mathrm{MIND}$ in the same range~\cite{Lundqvist2003} and $\beta$$\in$$[0, 1]$ is a weighting term.
The choice of $s$ is not trivial, as the magnitude of change per dissimilarity measure from initial to ideal knot displacements $D_\mathrm{init,ideal}(E_\mathrm{dissim}$) is unknown.
We tested 3 strategies, namely (i) using a fixed parameter $s$, (ii) using the initial gradient magnitude via 
\begin{equation}
s=\frac{D_\mathrm{init,ideal}(E_\mathrm{NMI})}{D_\mathrm{init,ideal}(E_\mathrm{MIND})} \approx 
\frac{\| \nabla E_\mathrm{NMI}(\mathbf{I}_\mathrm{f},\mathbf{I}_\mathrm{m}(d(\mathbf{k}_\mathrm{init,q}))) \|_{2}}{\| \nabla E_\mathrm{MIND}(\mathbf{I}_\mathrm{f},\mathbf{I}_\mathrm{m}(d(\mathbf{k}_\mathrm{init,q}))) \|_{2}}
\label{eq:grad}
\end{equation}
or (iii) basing it on the change in dissimilarity during registration: 
\begin{equation}
s=
\frac{D_\mathrm{init,ideal}(E_\mathrm{NMI})}
{D_\mathrm{init,ideal}(E_\mathrm{MIND})} 
\approx 
\frac{| E_\mathrm{NMI}(\mathbf{I}_\mathrm{f},\mathbf{I}_\mathrm{m}) - E_\mathrm{NMI}(\mathbf{I}_\mathrm{f},\mathbf{I}_\mathrm{m}(d(\mathbf{k}_\mathrm{init,q}))) |}
{| E_\mathrm{MIND}(\mathbf{I}_\mathrm{f},\mathbf{I}_\mathrm{m}) - E_\mathrm{MIND}(\mathbf{I}_\mathrm{f},\mathbf{I}_\mathrm{m}(d(\mathbf{k}_\mathrm{init,q}))) |}.
\end{equation}
The final cost function is
$F(d(\mathbf{k})) = 
E_\mathrm{dissim}(\mathbf{I}_\mathrm{f},\mathbf{I}_\mathrm{m}(d(\mathbf{k}))) + 
\lambda \, R(\mathbf{k})$
where $R(\mathbf{k})$ regularizes the displacements at the control points by their TV or L2 norm.

\section{Experiments and Results}

\subsection{Image Synthesis}

Images intensities were linearly scaled to [0,255].
Suitable image regions were cropped to fit the network size instead of resizing the image in-plane~\cite{wolterink2017deep}, as resizing lead to distortions in the synthesized images due to systematic differences in image size between the two modalities.
Image regions (ROIs) of size 286$\times$286$\times$$C$, for $C$$\in$$\{3,12\}$ were extracted randomly from the training data.
Dark ROIs, with a mean intensity of less than 1/4 of that of the ROI in the center of the 3D image, were not selected to avoid including a lot of background.
ROIs were further randomly cropped to 256$\times$256$\times$$C$ during network training.
A Cycle-GANs was trained for 200 epochs per image region (thorax or abdomen).
We used a training regime as previously reported~\cite{wolterink2017deep}, namely Adam optimizer, learning rate fixed to 0.0002 for 1-100 epochs and linearly reduced to 0 for 101-200 epochs, $\lambda_\mathrm{cyc}$$=$$10$.
3D test images were created using an in-plane stride length of
$S$$=$$4$ and a channel stride length of 
$S_{C}$$=$$2$ for $C$$=$$3$ and
$S_{C}$$=$$4$ for $C$$=$$12$.

\subsection{Image Registration}

MR-CT deformable image registration based on image multi-modal similarity measure MIND, NMI, or NMI+MIND 
was compared with image synthesis and then deformable registration using local NCC as image similarity. All registrations used ourDIR framework with total variation regularization.
Registration parameters were optimized via grid search. These  are the weighting of regularization term $\lambda$$\in$$\{0.0125, 0.025, 0.05, 0.1, 0.2\}$, 
the control point spacing $s$$\in$$\{8, 10, 12, 14,16\}$ pixels, and 
the number of multi-resolution levels $l$$\in$$\{2, 3, 4\}$. The best strategy for combining NMI and MIND was using the initial gradient magnitude, i.e. Eq.~(\ref{eq:grad}), and $\beta$$=$$0.8$.

\subsection{Results}

\begin{figure}[tbh!]
\includegraphics[trim = 40mm 10mm 35mm 10mm, clip, height=0.15\textheight]{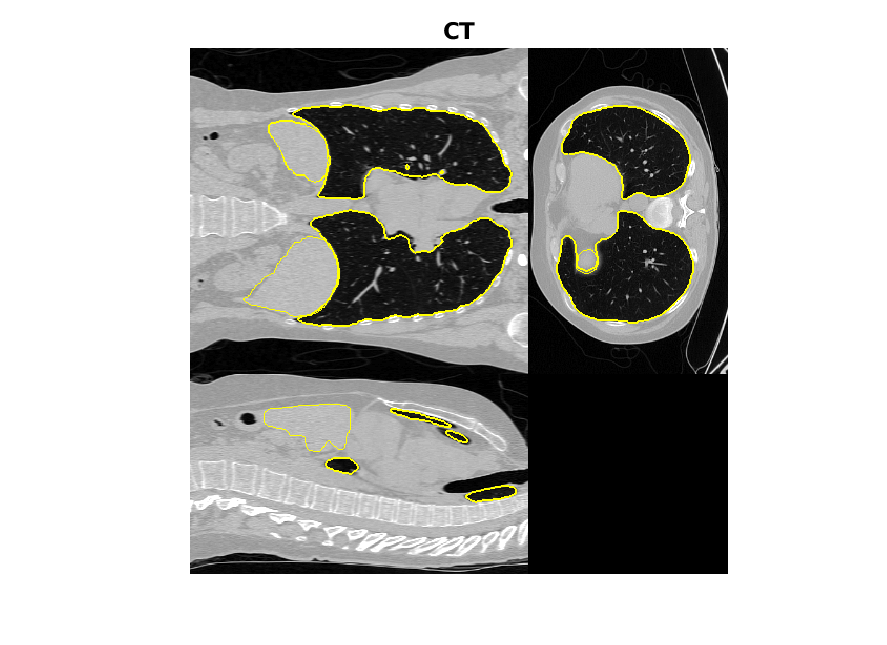} 
\includegraphics[trim = 40mm 10mm 35mm 10mm, clip, height=0.15\textheight]{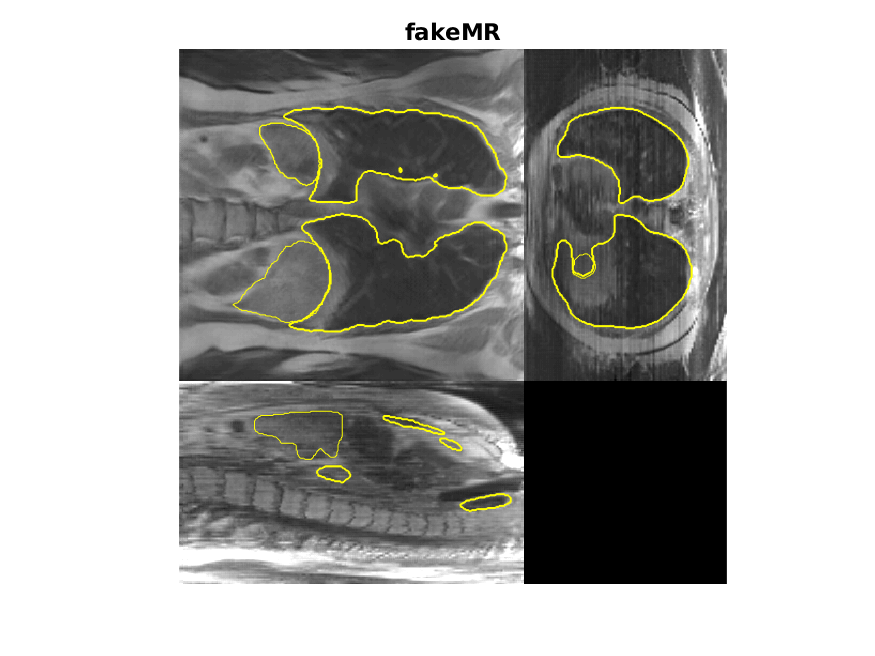} 
\includegraphics[trim = 40mm 10mm 35mm 10mm, clip, height=0.15\textheight]{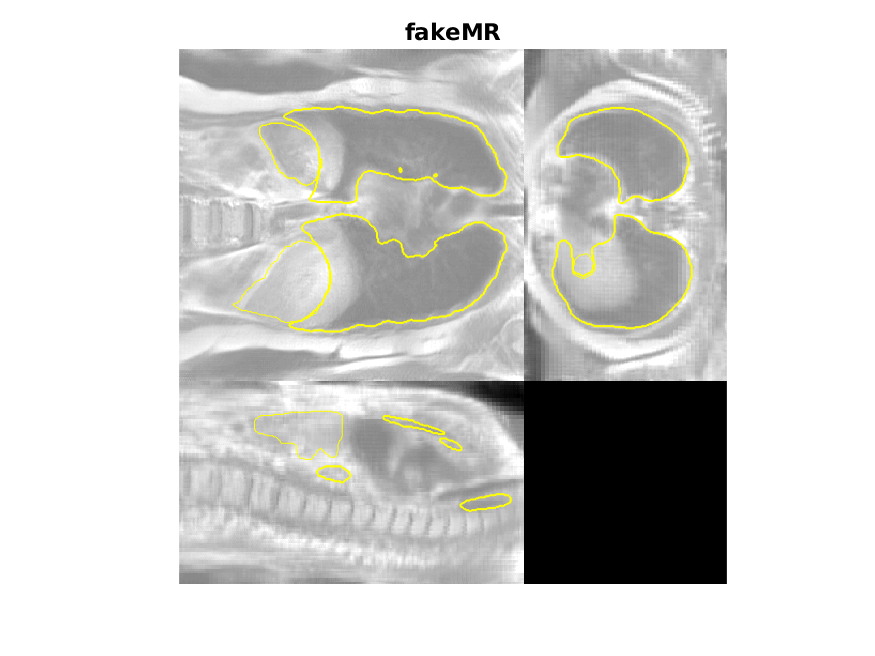} 
\includegraphics[trim = 40mm 10mm 35mm 10mm, clip, height=0.15\textheight]{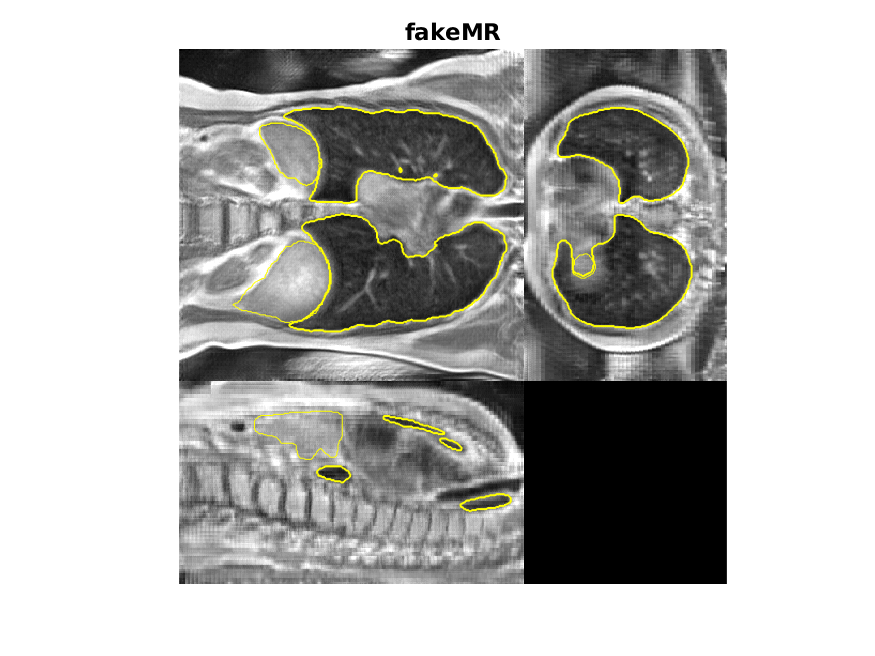} 
\includegraphics[trim = 40mm 10mm 35mm 10mm, clip, height=0.15\textheight]{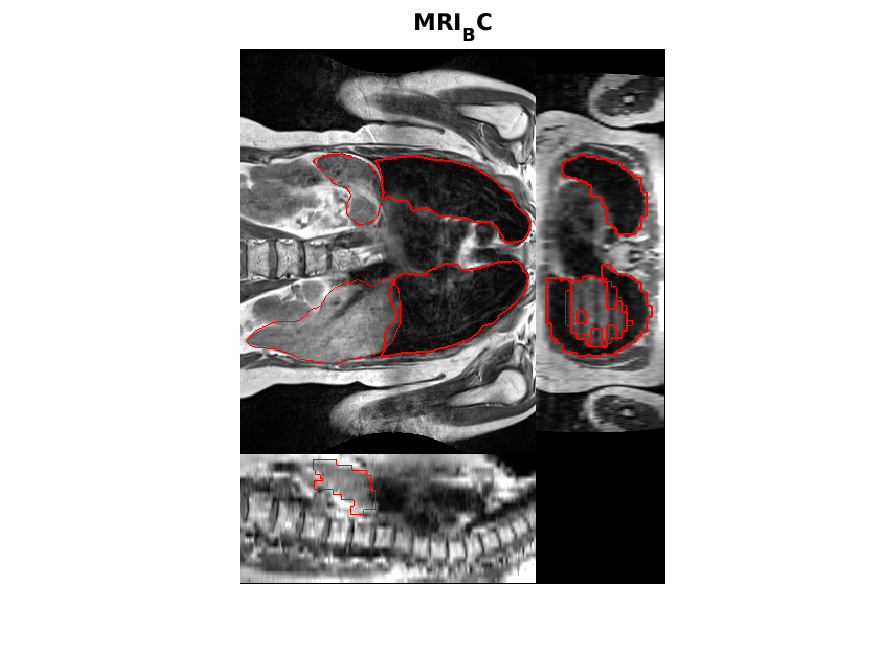}
\\
\includegraphics[trim = 40mm 10mm 35mm 10mm, clip, height=0.15\textheight]{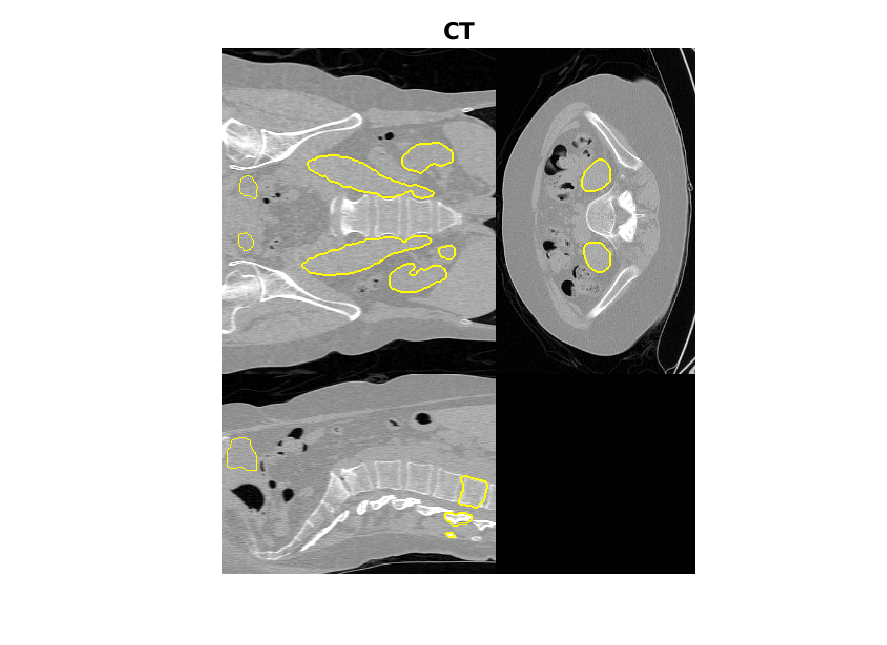} 
\includegraphics[trim = 40mm 10mm 35mm 10mm, clip, height=0.15\textheight]{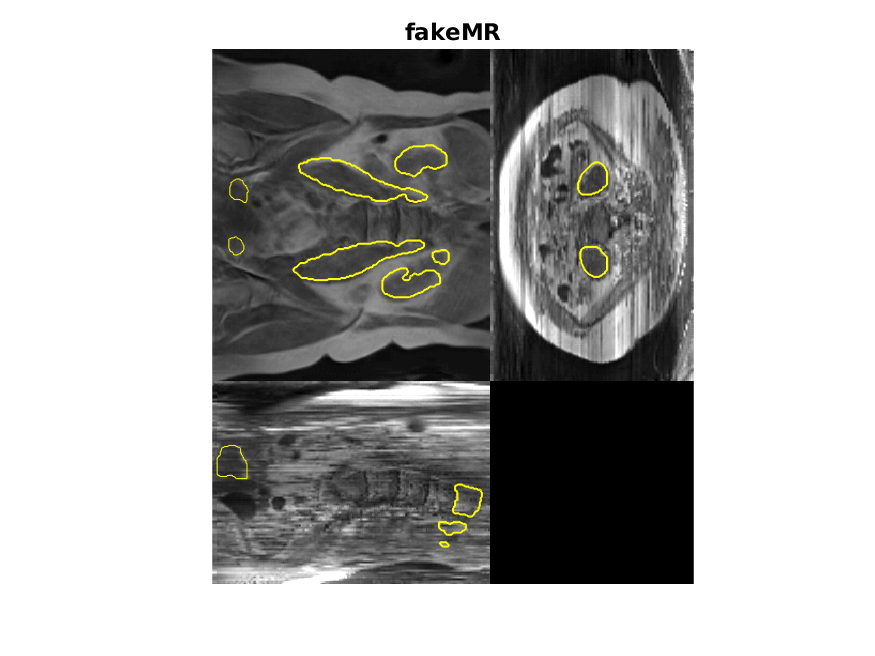} 
\includegraphics[trim = 40mm 10mm 35mm 10mm, clip, height=0.15\textheight]{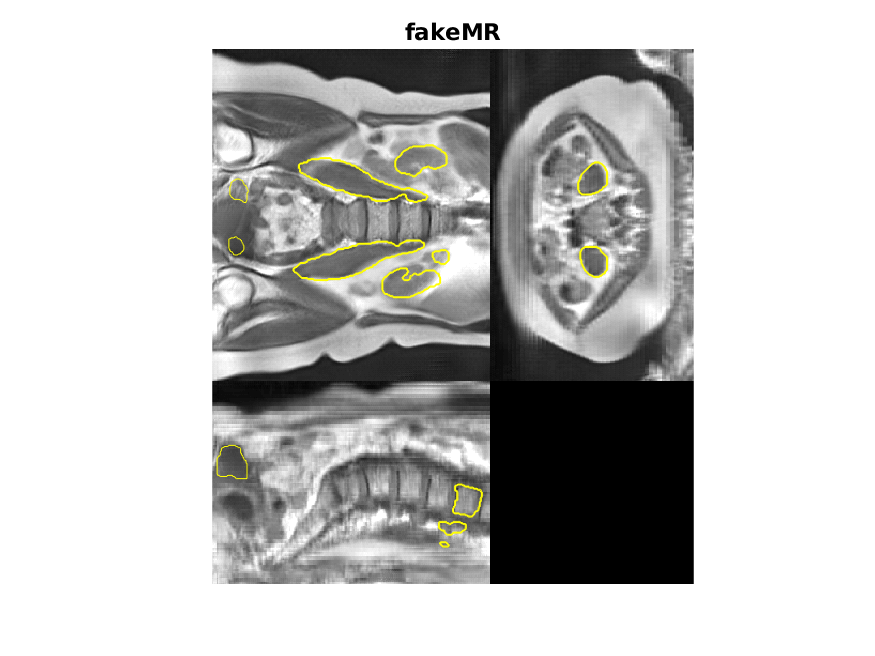} 
\includegraphics[trim = 40mm 10mm 35mm 10mm, clip, height=0.15\textheight]{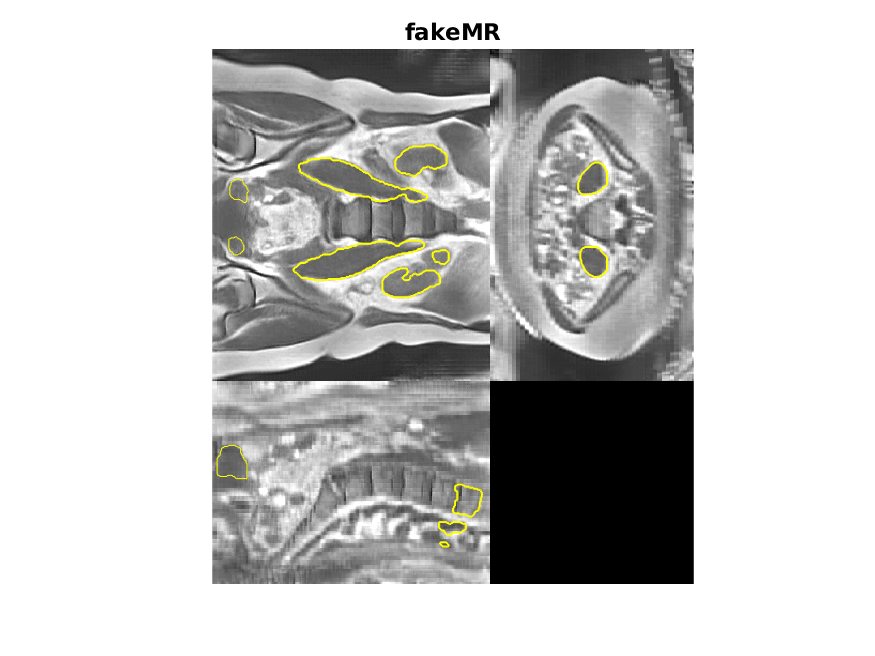} 
\includegraphics[trim = 40mm 10mm 35mm 10mm, clip, height=0.15\textheight]{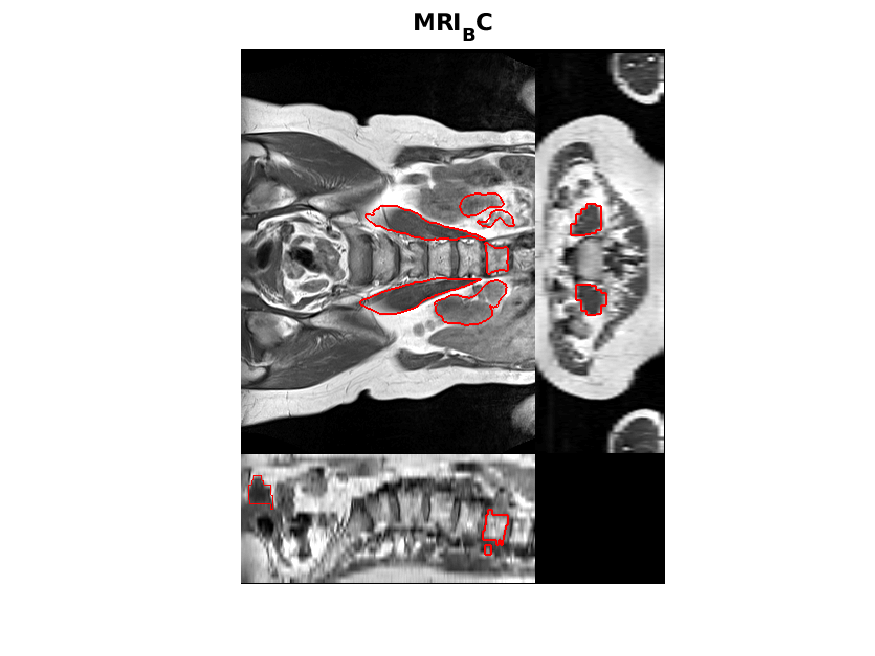}\\
\caption{Illustration of MR synthesis from CT for (top) thoracic and (bottom) abdominal region. (a) original CT,  (b-d) synthesized MRIs from (b)  256$^2$x3 ROIs, (c)  256$^2$x12 ROIs, (d)  256$^2$x12 ROIs and 34$\times$34 patches, (e) original MRI. Original MR (CT) contours in red (yellow). }
\label{fig:fakeMR}
\end{figure}

\begin{figure}[tbh!]
\centering
\includegraphics[trim = 40mm 10mm 35mm 10mm, clip, height=0.15\textheight]{resFigs/10000067_MRI_BC_superior_origContour.png} 
\includegraphics[trim = 40mm 10mm 35mm 10mm, clip, height=0.15\textheight]{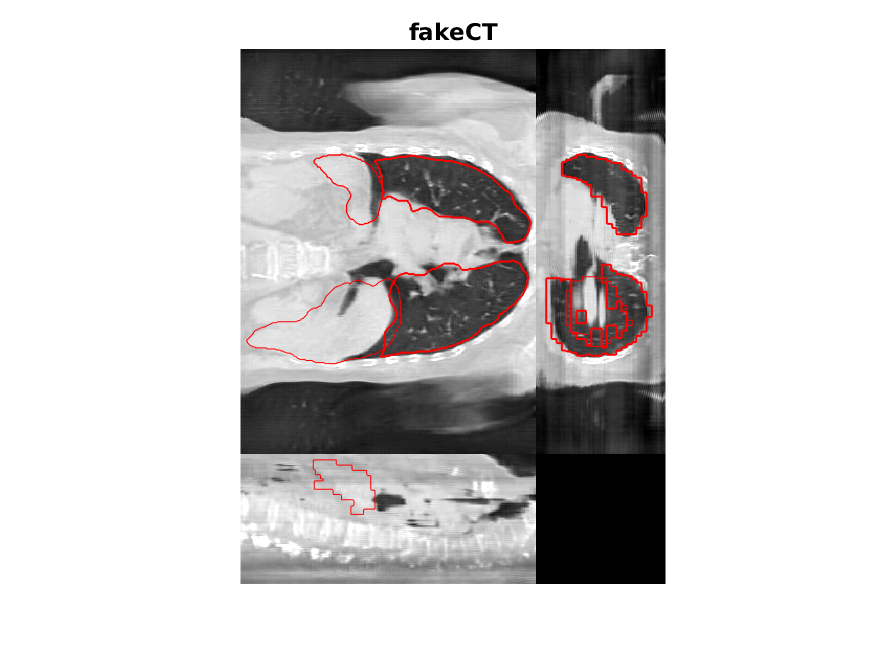} 
\includegraphics[trim = 40mm 10mm 35mm 10mm, clip, height=0.15\textheight]{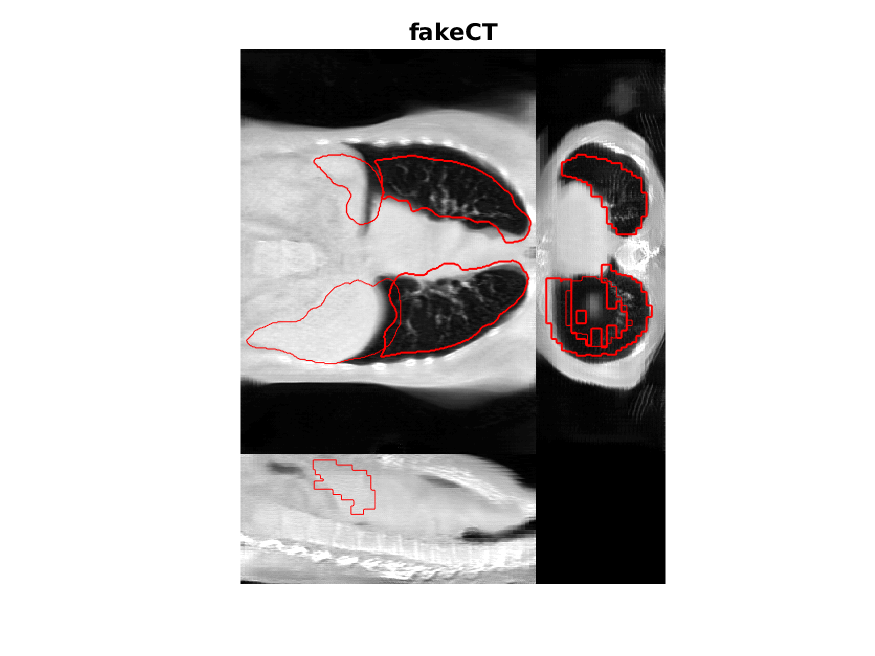}
\includegraphics[trim = 40mm 10mm 35mm 10mm, clip, height=0.15\textheight]{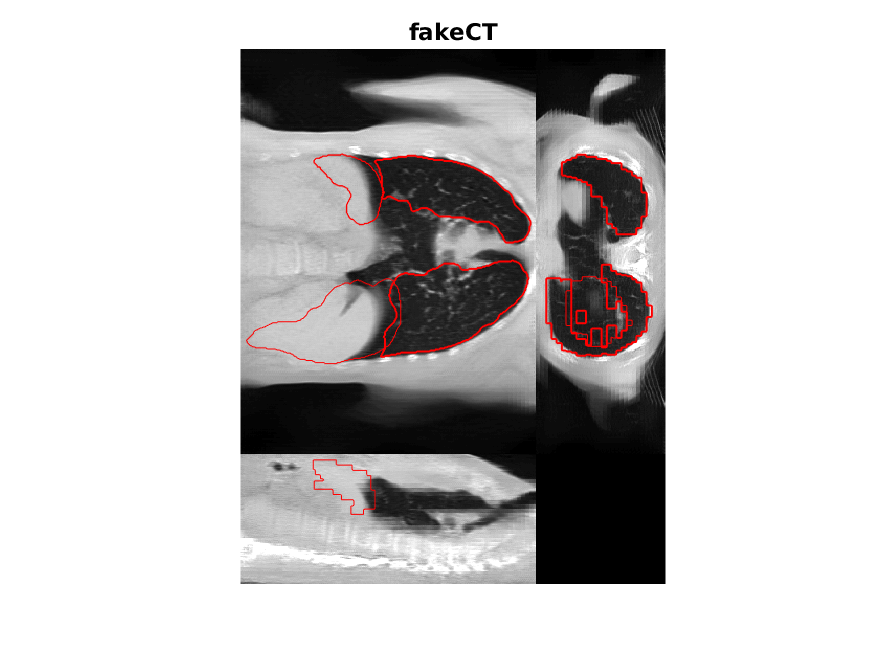}
\includegraphics[trim = 40mm 10mm 35mm 10mm, clip, height=0.15\textheight]{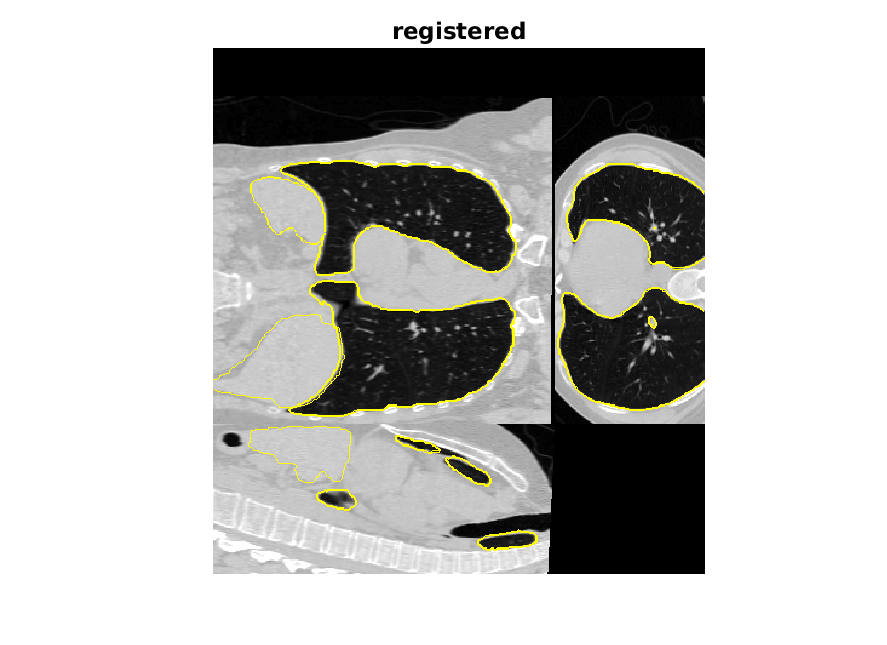}
\\
\includegraphics[trim = 40mm 10mm 35mm 10mm, clip, height=0.15\textheight]{resFigs/10000067_MRI_BC_inferior_origContour.png}
\includegraphics[trim = 40mm 10mm 35mm 10mm, clip, height=0.15\textheight]{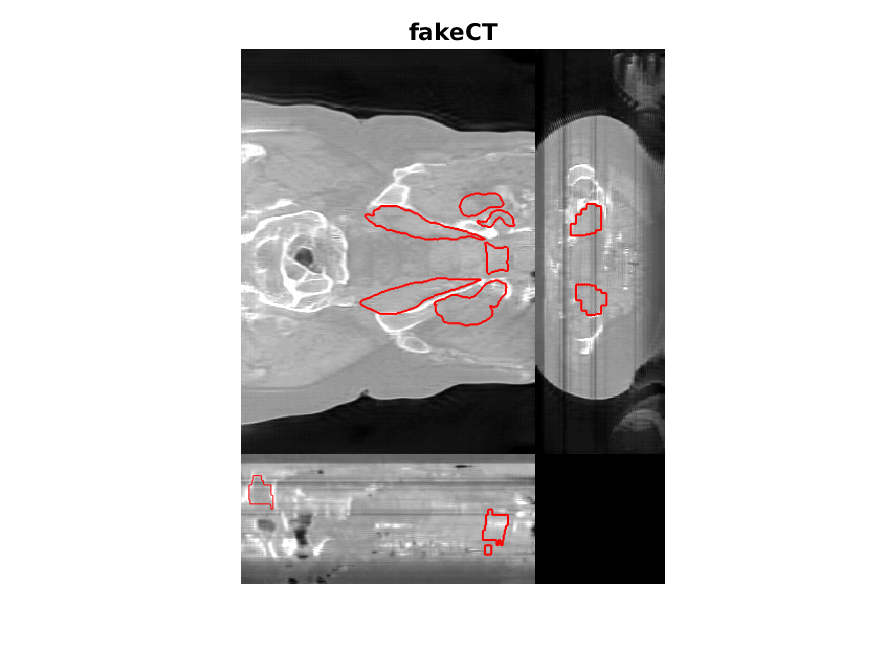}
\includegraphics[trim = 40mm 10mm 35mm 10mm, clip, height=0.15\textheight]{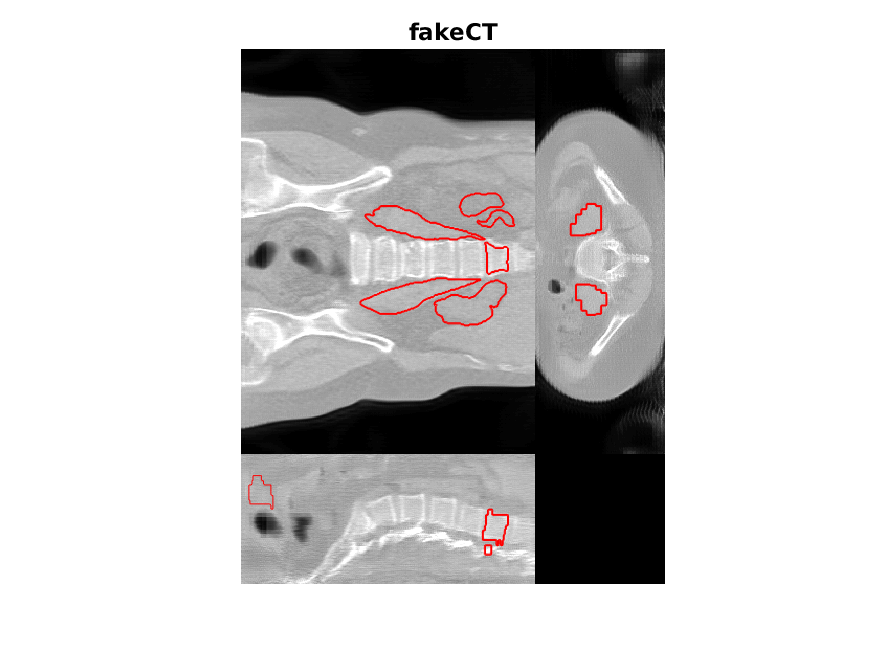}
\includegraphics[trim = 40mm 10mm 35mm 10mm, clip, height=0.15\textheight]{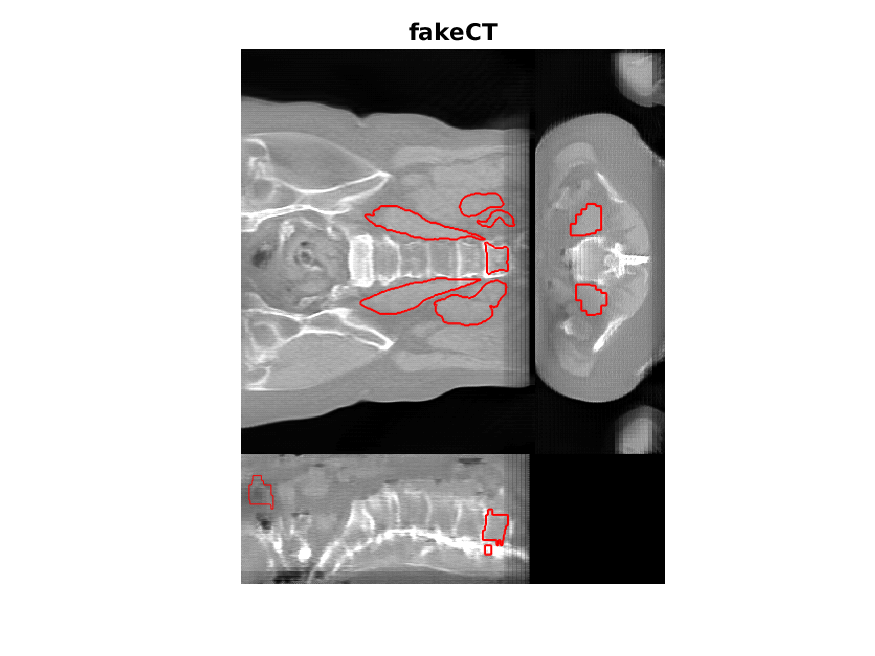}
\includegraphics[trim = 40mm 10mm 35mm 10mm, clip, height=0.15\textheight]{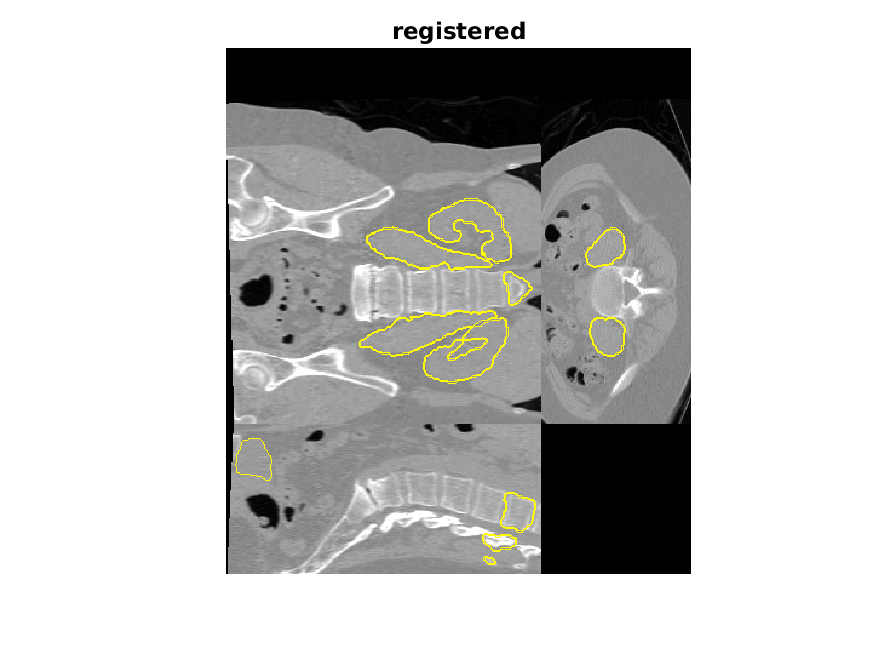}
\caption{Illustration of CT synthesis from MR for (top) thoracic and (bottom) abdominal region. (a) original MRI, (b-d) synthesized CTs from (b) 256$^2$x3 ROIs, (c)  256$^2$x12 ROIs, (d) 256$^2$x12 ROIs and 34$\times$34 patches, (e) original CT rigidly aligned. Original MR (CT) contours in red (yellow).}
\label{fig:fakeCT}
\end{figure}

Example of synthesized images are shown in Figs.~\ref{fig:fakeMR}-\ref{fig:fakeCT}. 
Inconsistency across slices can be seen when ROIs with few slices ($C$=$3$) are used. 
Synthesized image structures do mostly not adhere to the contours of the lung segmentations from the real image. To be realistic, the generators had to learned the substantial bias in lung volume between the two modalities, see Table~\ref{tab:volumeChange}. CT images were generally acquired in end-inhale state, while MRIs in end-exhale. Changing the patchGAN discriminator to a shallower architecture, such that each output node has a  receptor field of $P$$=$34$\times$34 instead of 70$\times$70 could sometimes reduced the misalignment of the lung segmentations (e.g. Figs.~\ref{fig:fakeMR}d), but was less powerful in image synthesis (e.g. region between the lungs in Fig.\ref{fig:fakeCT}d). 

\begin{figure}[tbh!]
\centering
\begin{tabular}{cccccccc}
\includegraphics[trim = 40mm 10mm 35mm 10mm, clip, height=0.126\textheight]{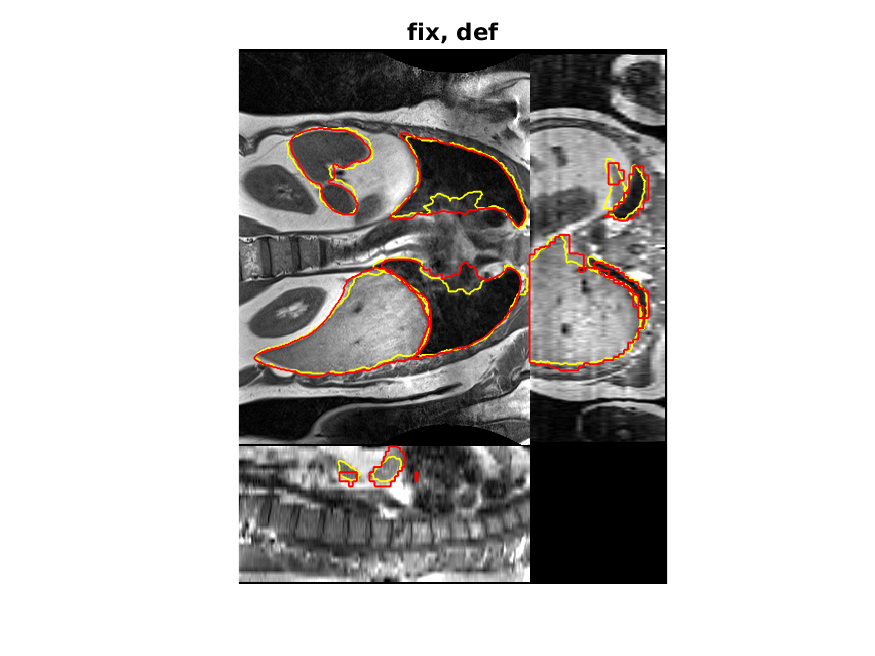} &
\includegraphics[trim = 40mm 10mm 35mm 10mm, clip, height=0.126\textheight]{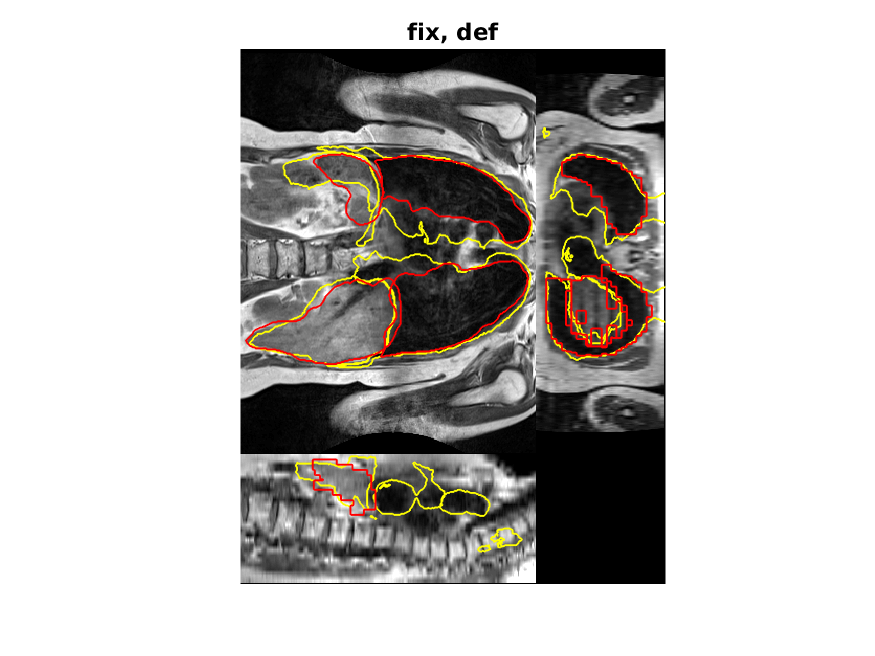} &
\includegraphics[trim = 40mm 10mm 35mm 10mm, clip, height=0.126\textheight]{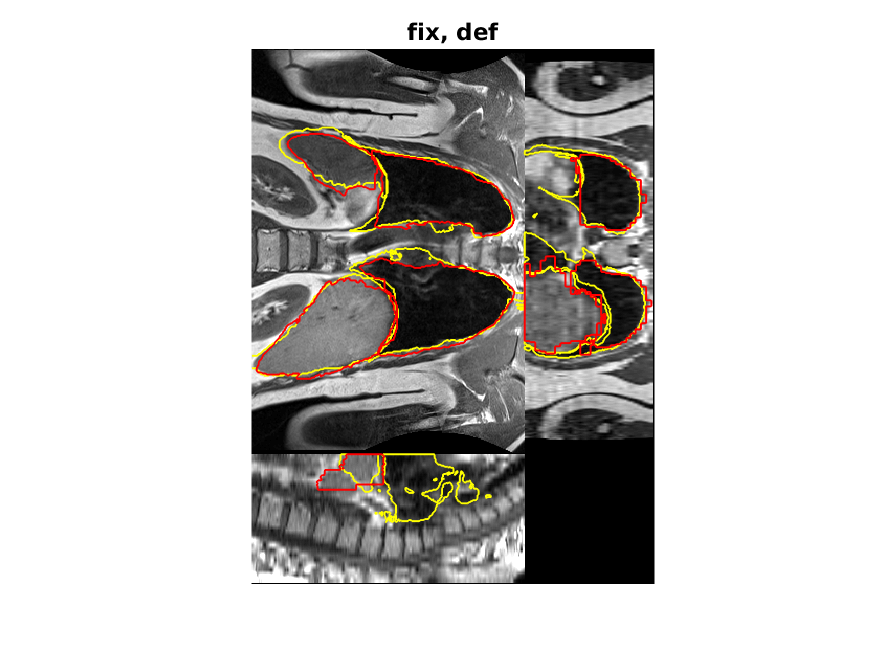} &
\includegraphics[trim = 40mm 10mm 35mm 10mm, clip, height=0.126\textheight]{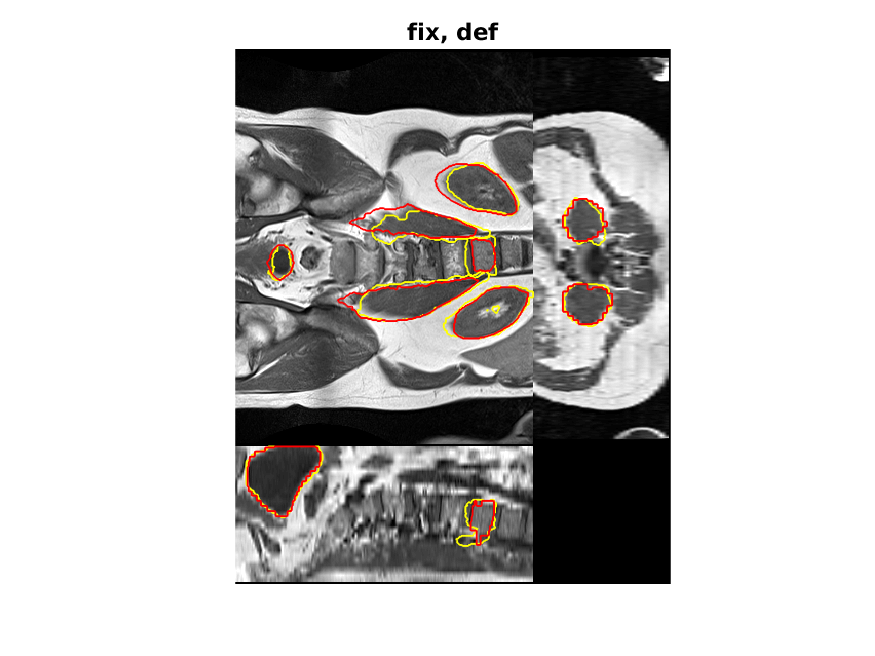} &
\includegraphics[trim = 40mm 10mm 35mm 10mm, clip, height=0.126\textheight]{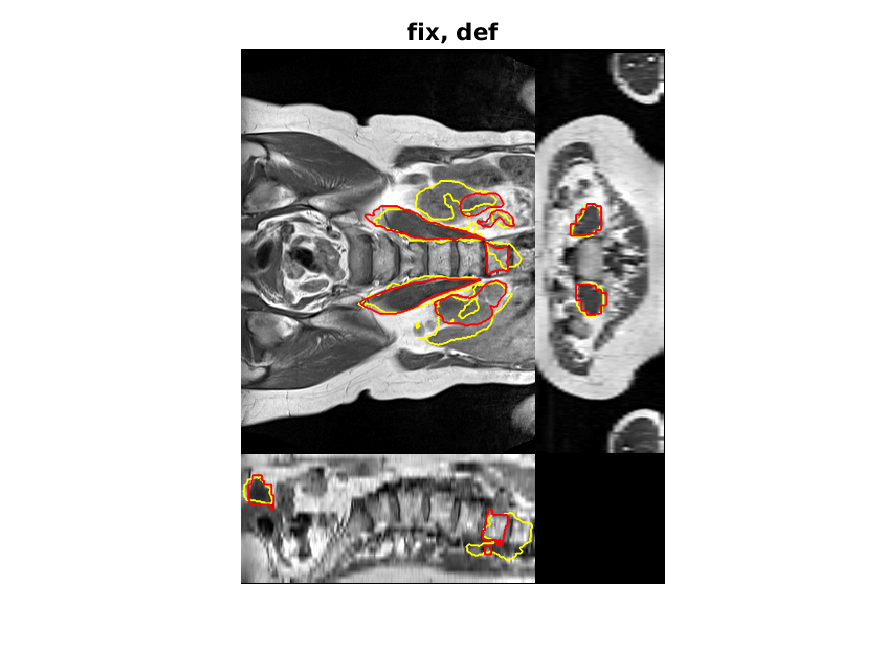} &
\includegraphics[trim = 40mm 10mm 35mm 10mm, clip, height=0.126\textheight]{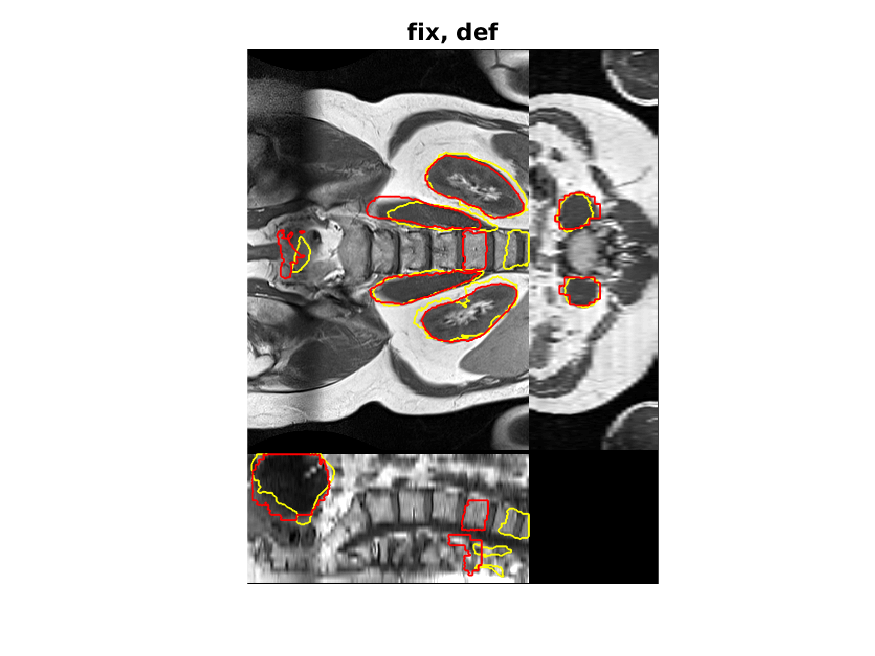} &
\\
86.7\% & 59.8\% & 77.0\% & 82.0\% & 70.9\% & 83.8\%\\
\includegraphics[trim = 40mm 10mm 35mm 10mm, clip, height=0.126\textheight]{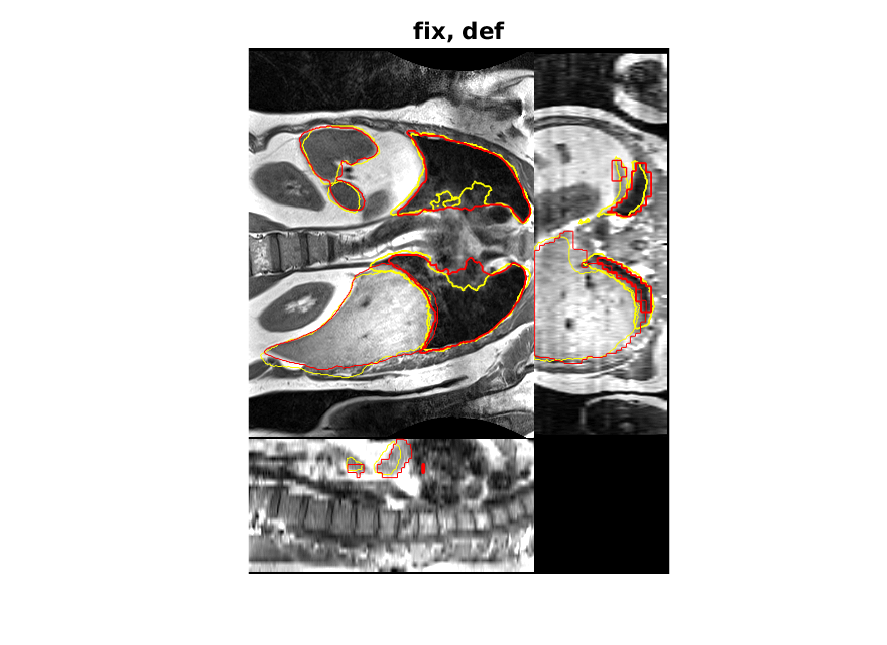} &
\includegraphics[trim = 40mm 10mm 35mm 10mm, clip, height=0.126\textheight]{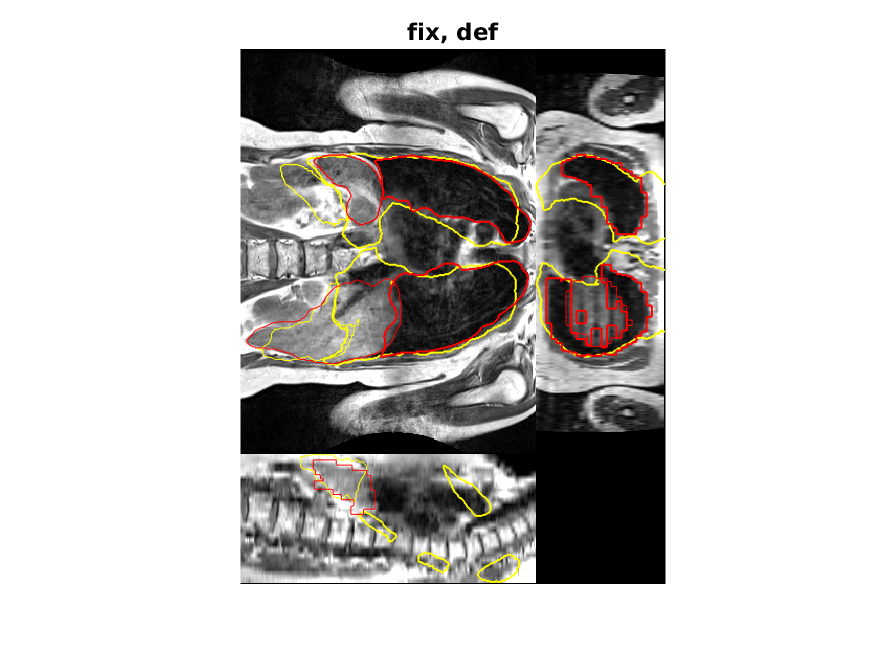} &
\includegraphics[trim = 40mm 10mm 35mm 10mm, clip, height=0.126\textheight]{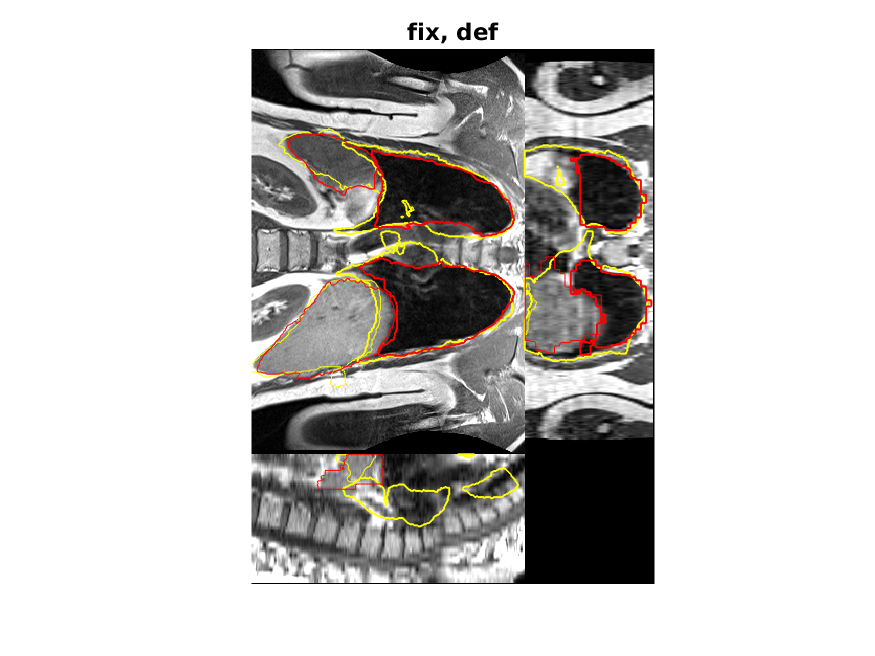} &
\includegraphics[trim = 40mm 10mm 35mm 10mm, clip, height=0.126\textheight]{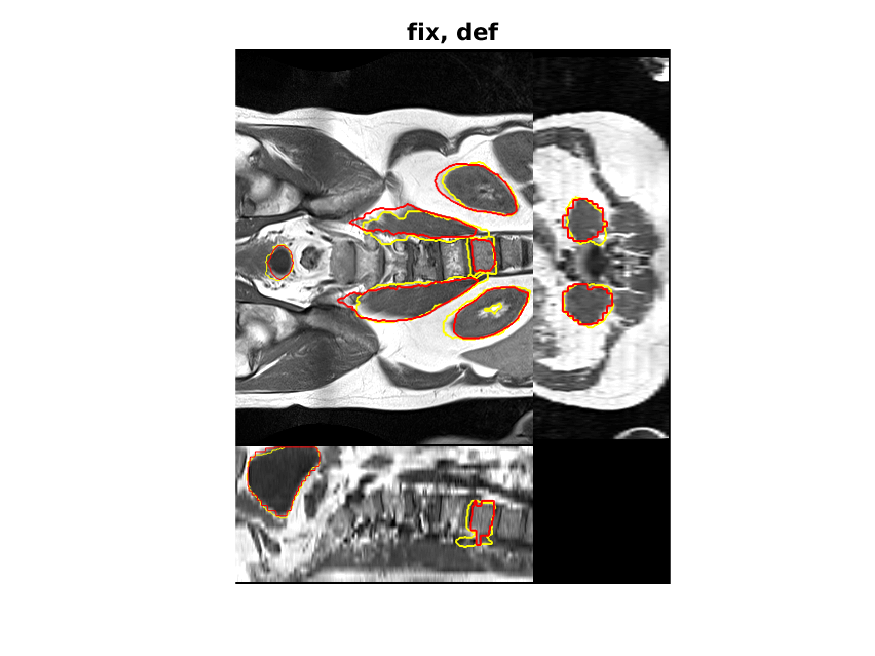} &
\includegraphics[trim = 40mm 10mm 35mm 10mm, clip, height=0.126\textheight]{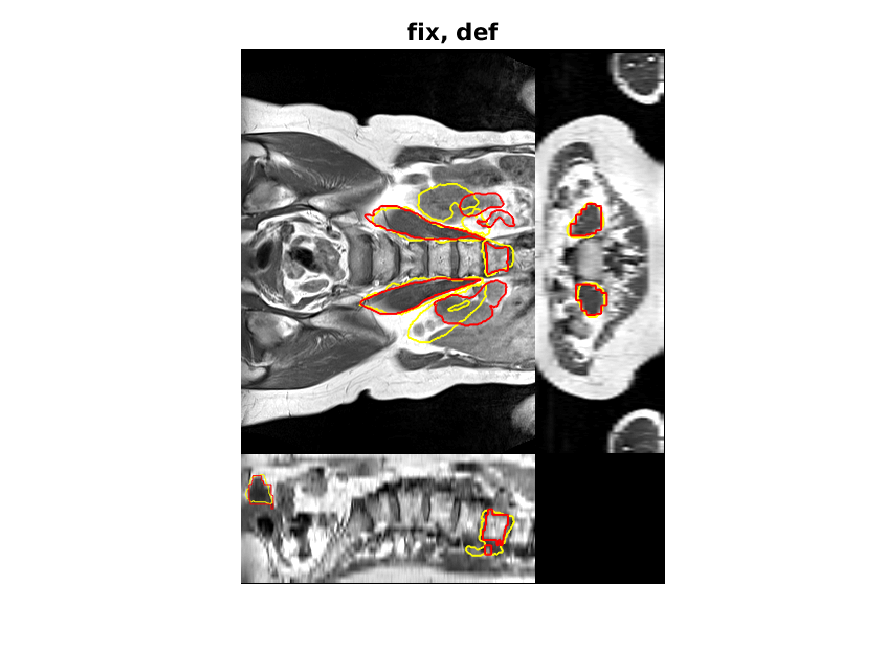} &
\includegraphics[trim = 40mm 10mm 35mm 10mm, clip, height=0.126\textheight]{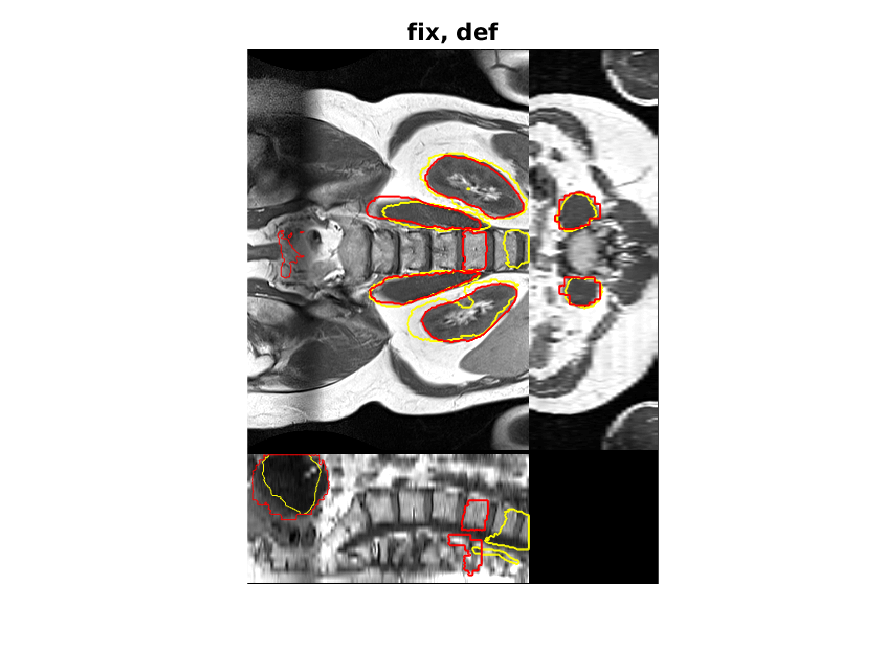} &
\\
76.3\% & 35.3\% & 71.2\% & 83.2\% & 67.8\% & 80.9\%\\
\end{tabular}
\caption{Deformable image registration results for (top) CT-to-MRI based on NMI, (bottom) synthesizedMRI-to-MRI based on local NCC (256$^2$x12 ROIs, 34$\times$34 patches). Image: original MR, yellow contour: original MR, red contour: deformed CT or synthesized MRI, value: mean Dice. }
\label{fig:registrationResults}
\end{figure}

The performance of the deformable image registration for the original images (CT, MR) and the cycle-GAN synthesized images are listed in~Table~\ref{tab:pTVDICE}. 
While synthesized images can achieve a similar overlap than multi-modal NMI registration for the abdomen (77.4 vs. 78.8\%), they are substantially worse for the thorax due to the bias in lung volume. 
Performance of synthesized CT images, which were more affected by synthesized volume lung volume changes, was generally lower than for synthesized MRIs. Results are shown in Fig.~\ref{fig:registrationResults}.

\begin{table}[tbh!]
\centering
\begin{tabular}{|l|rrrrr|rrrrrr|rrrrrrrrrrr}
\hline
 & \rotatebox{40}{Liver} & \rotatebox{40}{Spleen} & \rotatebox{40}{Gallb.} & \rotatebox{40}{rLung} & \rotatebox{40}{lLung} & \rotatebox{40}{Bladder} & \rotatebox{40}{lVert1} & \rotatebox{40}{rKidney} & \rotatebox{40}{lKidney} & \rotatebox{40}{rPsoas} & \rotatebox{40}{lPsoas}\\
\hline
meanCT & 1896  & 244   & 42 & 2598 & 2253 & 198  & 63 & 185 & 197 & 208 & 188 \\
meanMR & 1576  & 248  & 201 & 1338 & 1144 & 153  & 62 & 211 & 225 & 158 & 180  \\
\hline
Ratio (\%) & 120 & 98 & 21 & 194 & 197 & 129 & 101 & 88 & 88 & 131 & 105 \\
\hline
\end{tabular}
\caption{Mean volume of segmented structures in cm$^3$ per modality and ratio meanCT/meanMR for unpaired training data.}
\label{tab:volumeChange}
\end{table}
\begin{table}[tbh!]
\centering
\begin{tabular}{|l|rrrr|rrrr|rrrr|r|r|r|r|r|r|rrrrrrrrrrr} 
\hline
        & \multicolumn{4}{c|}{CT-MR} &  \multicolumn{4}{c|}{Synthesized MR}& \multicolumn{4}{c|}{Synthesized CT}\\
        &        &      &     &  NMI+ & $C$=3  & $C$=12 & $C$=12 & $C$=12 & $C$=3 & $C$=12 & $C$=12 & $C$=12 \\
        &  rigid & MIND & NMI &  MIND & $P$=70 & $P$=70 & $P$=34 & $P$=22 & $P$=70 & $P$=70 & $P$=34 & $P$=22 \\
\hline
Thorax  & 55.2 & 65.4 & \bf{75.2} & \underline{\bf{75.7}} & 65.2 & 62.2 & 62.3 & 55.4 & 58.1 & 59.3 & 54.6 & 55.2\\
Abdomen & 60.6 & 73.9 & \underline{\bf{78.8}} & \bf{78.3} & 66.6 & \bf{76.9} & \bf{77.4} & 58.0 & 48.4 & 72.3 & 72.6 & 60.4 \\
Both    & 59.6 & 67.7 & \bf{76.7} & \underline{\bf{76.9}} & 65.6 & 69.0 & 69.1 & 56.6 & 52.5 & 64.6 & 63.1 & 57.8\\
\hline
\end{tabular}
\caption{DIR performance measured by mean Dice overlap ratio (\%) for original images (CT-MR) or for cycle-GAN synthesized MR or CT images, using ROIs of size 256$\times$256$\times$$C$ with discriminator based on $P$$\times$$P$ patches. Results within 5\% to the \underline{\bf{best result}} are marked in \bf{bold}.} 
\label{tab:pTVDICE}
\end{table}

\section{Conclusion}

We combined two multi-modal image similarity measure (NMI, MIND) and observed a similar performance as when using only NMI, and in contrast to~\cite{Heinrich2012} no improvement of MIND over NMI.

We investigated the usefulness of a fully unsupervised MR-CT image modality synthesis method for deformable image registration of MR and CT images. 
Against the established multi-modal deformable registration methods, synthesizing images via cycle-GAN and then using a robust mono-modal image similarity measure achieved at best a similar performance. In particular one has to be careful to have collections of the two image modalities which are balanced, i.e. not biased for a modality, as such differences are readily synthesized by the cycle-GAN framework. Ensuring that synthesized images are truly in spatial correspondence with the source image would require incorporating a deformable image registration into the cycle-GAN.

\noindent {\bf Acknowledgments:} We thank the EU’s 7th Framework Program (Agreement No. 611889, TRANS-FUSIMO) for funding and acknowledge NVIDIA for GPU support.

\bibliographystyle{splncs}
\bibliography{CTMRIregistration}

\end{document}